\documentclass[letterpaper, 10 pt, conference]{ieeeconf}  

\IEEEoverridecommandlockouts                              

\usepackage{graphicx}
\usepackage{comment}
\usepackage{amsmath,amssymb} 
\usepackage{color}

\usepackage{subfigure}
\usepackage{mathrsfs}
\usepackage{booktabs}
\usepackage{multirow}
\usepackage{array}
\usepackage{comment}
\usepackage{caption}

\usepackage{mdwlist}
\usepackage{hyperref}

\newcommand{\PreserveBackslash}[1]{\let\temp=\\#1\let\\=\temp}
\newcolumntype{C}[1]{>{\PreserveBackslash\centering}p{#1}}
\newcolumntype{R}[1]{>{\PreserveBackslash\raggedleft}p{#1}}
\newcolumntype{L}[1]{>{\PreserveBackslash\raggedright}p{#1}}

\title{\LARGE \bf
Cirrus: A Long-range Bi-pattern LiDAR Dataset
}

\author{Albert Author$^{1}$ and Bernard D. Researcher$^{2}$
\thanks{*This work was not supported by any organization}
\thanks{$^{1}$Albert Author is with Faculty of Electrical Engineering, Mathematics and Computer Science,
        University of Twente, 7500 AE Enschede, The Netherlands
        {\tt\small albert.author@papercept.net}}%
\thanks{$^{2}$Bernard D. Researcheris with the Department of Electrical Engineering, Wright State University,
        Dayton, OH 45435, USA
        {\tt\small b.d.researcher@ieee.org}}%
}

\author{Ze Wang$^{1}$, Sihao Ding$^{2}$, Ying Li$^{2}$, Jonas Fenn$^2$, Sohini Roychowdhury$^2$, Andreas Wallin$^2$, Lane Martin$^3$, \\Scott Ryvola$^3$, Guillermo Sapiro$^4$, and Qiang Qiu$^1$
	\thanks{$^{1}$ Ze Wang and Qiang Qiu are with Purdue University, USA. {\tt\small \{zewang, qqiu\}@purdue.edu}}
	\thanks{$^{2}$ Sihao Ding, Ying Li, Jonas Fenn, Sohini Roychowdhury, and Andreas Wallin are with Volvo Cars Technology, USA. {\tt\small \{sihao.ding, ying.li.5, jonas.fenn, sohini.roy.chowdhury, andreas.wallin1\}@volvocars.com}}
	\thanks{$^3$ Lane Martin and Scott Ryvola are with Luminar Technologies, Inc., USA. {\tt\small \{lane, scott.ryvola\}@luminartech.com}}
	\thanks{$^4$ Guillermo Sapiro is with Duke University, USA.  {\tt\small guillermo.sapiro@duke.edu}}
}

\begin{document}

\maketitle
\thispagestyle{empty}
\pagestyle{empty}

\begin{abstract}

In this paper, we introduce Cirrus, a new long-range bi-pattern LiDAR public dataset for autonomous driving tasks such as 3D object detection, critical to highway driving and timely decision making. Our platform is equipped with a high-resolution video camera and a pair of LiDAR sensors with a 250-meter effective range, which is significantly longer than existing public datasets. We record paired point clouds simultaneously using both Gaussian and uniform scanning patterns. 
Point density varies significantly across such a long range, and different scanning patterns further diversify object representation in LiDAR. 
In Cirrus, eight categories of objects are exhaustively annotated in the LiDAR point clouds for the entire effective range.
To illustrate the kind of studies supported by this new dataset, we introduce LiDAR model adaptation across different ranges, scanning patterns, and sensor devices.
Promising results show the great potential of this new dataset to the robotics and computer vision communities.

\end{abstract}

\section{Introduction}

Deep learning has significantly improved robustness and efficiency of autonomous driving in terms of visual perception, planning, and mapping.
Public datasets play a crucial role here.
For example, being the first large scale autonomous driving public dataset, KITTI \cite{geiger2013vision} has been widely adopted for developing and evaluating many state-of-the-art autonomous driving algorithms, or their key components such as object recognition, in the past several years. 
However, it remains unconfirmed if these algorithms generalize well to unseen scenarios, e.g., with different scanning patterns or range, mostly due to the lack of corresponding datasets for validation.

\begin{figure}[]
	\centering
	\includegraphics[width=\linewidth]{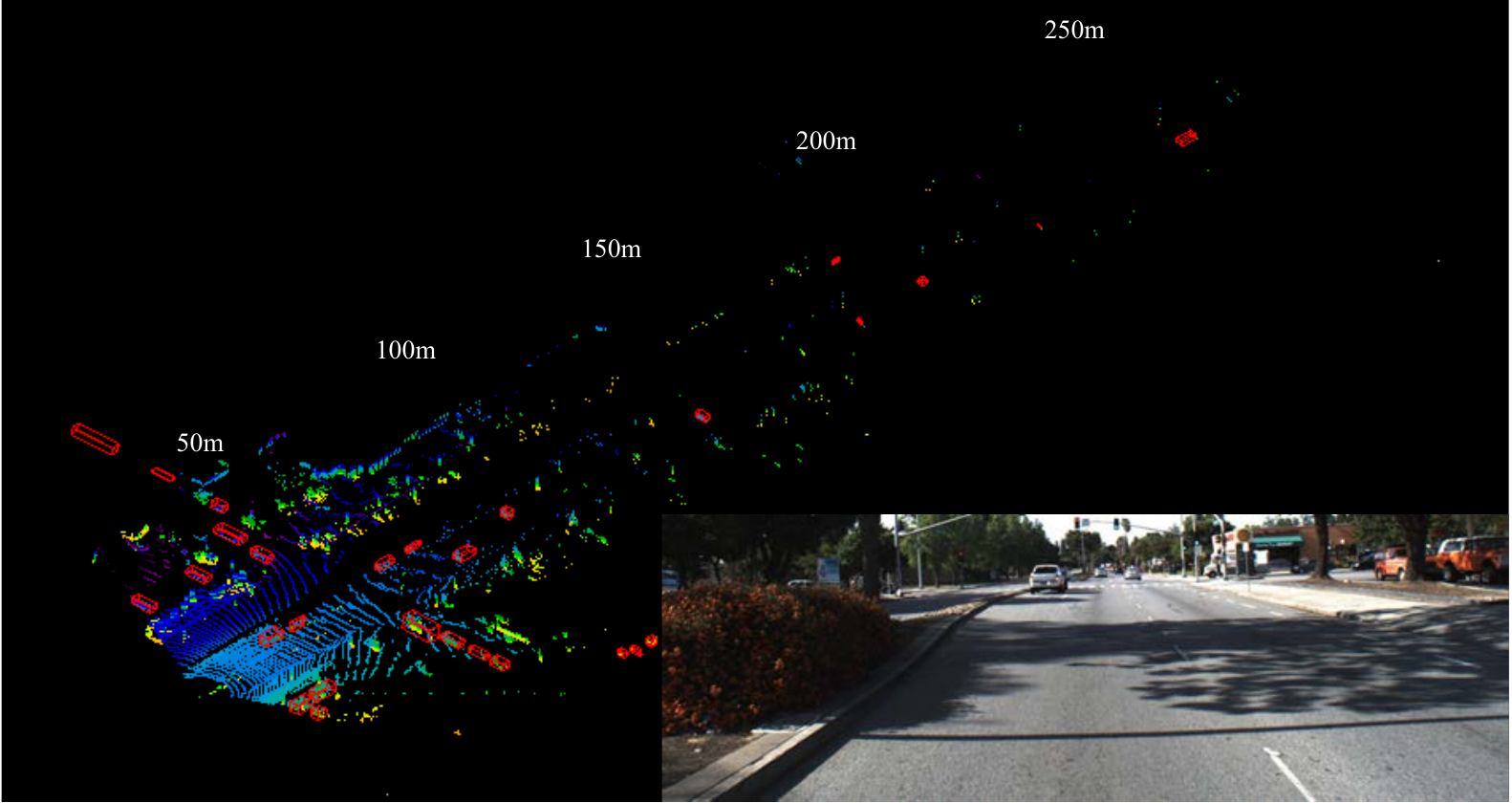}\\
	\vspace{1mm}
	\includegraphics[width=\linewidth]{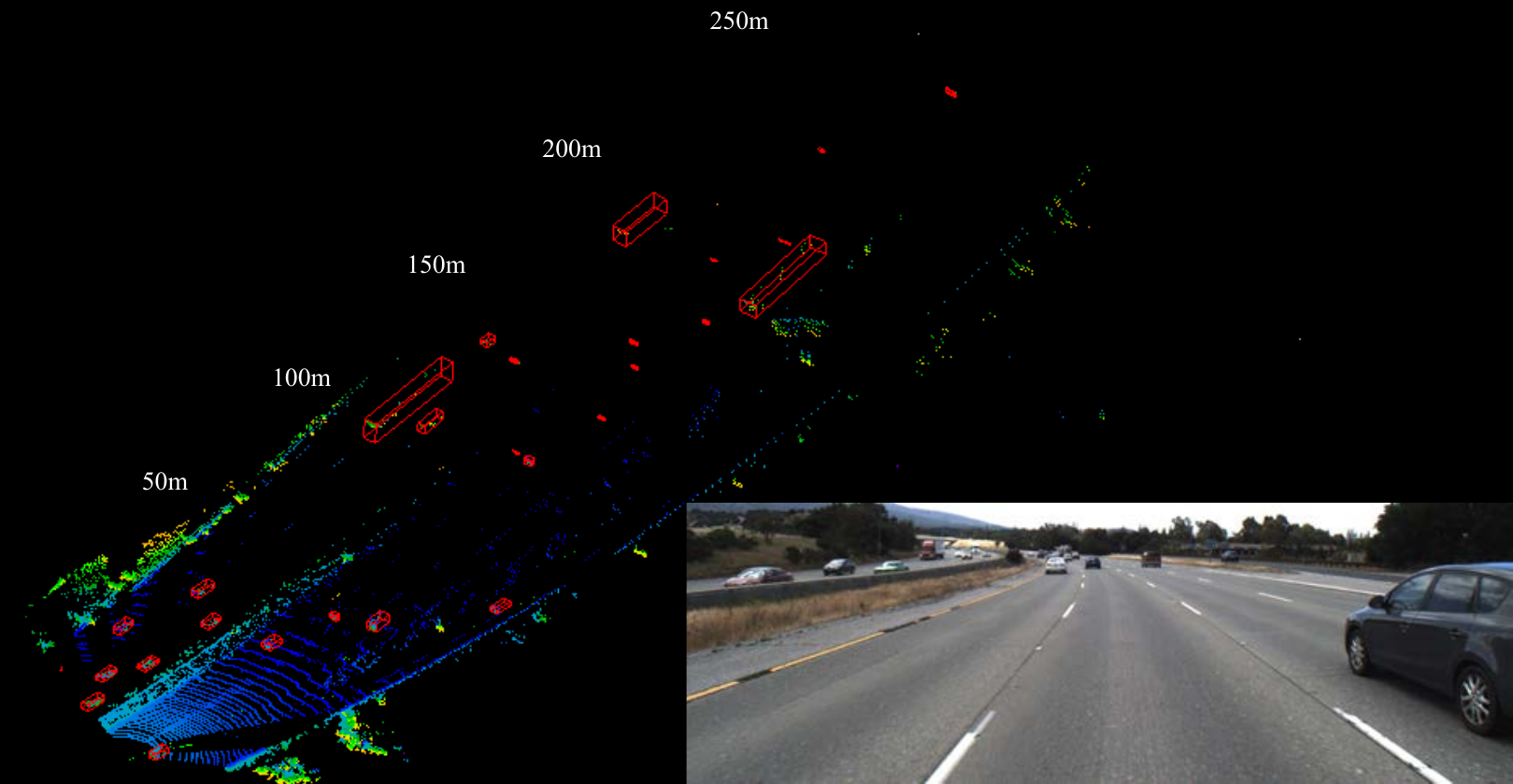}\\
	\caption{Example LiDAR point clouds from the Cirrus dataset with bounding boxes. Distance is marked in white.
	}
	\label{fig:main} 
\end{figure}

\begin{figure*}[t]
	\centering
	\includegraphics[width=\linewidth]{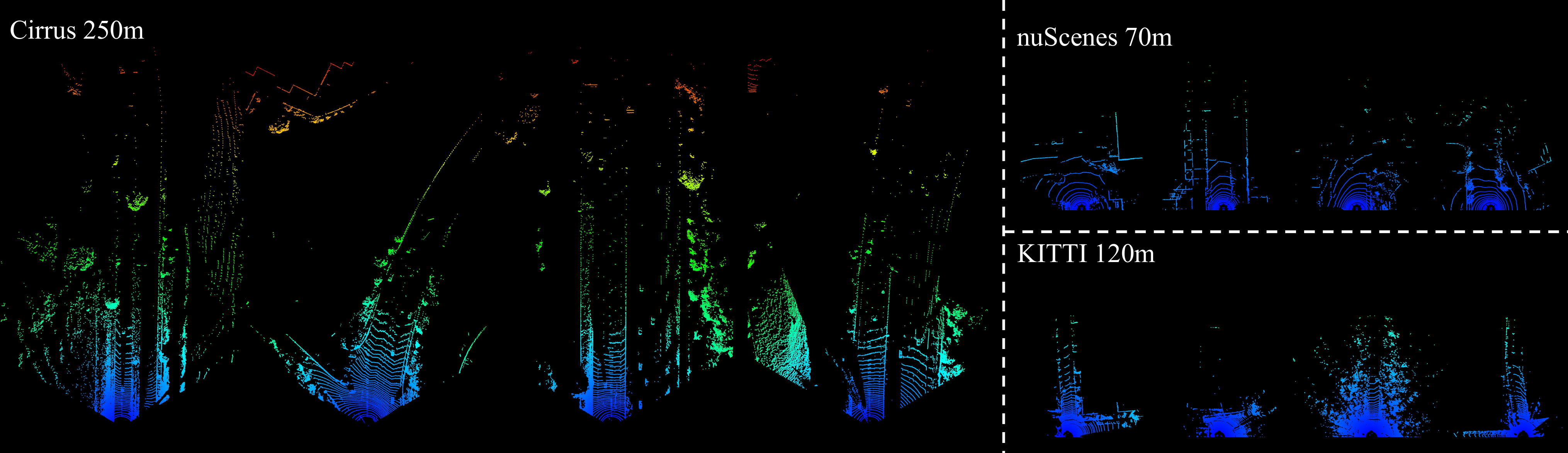}\\
	\caption{Point cloud samples from different datasets. Warmer colors indicate longer point distances. The respective effective ranges are marked. When cars drive at 75mph, the 120m, 70m, and 250m effective ranges allow 3.5s, 2s and 7.5s reaction time, respectively.
	}
	\label{fig:effective} 
\end{figure*}

In this paper, we introduce Cirrus, a new long-range bi-pattern LiDAR dataset for autonomous driving tasks.
Cirrus is developed to enhance existing public LiDAR datasets with additional diversity in terms of sensor model, range, and scanning pattern. 
A long effective range allows object detection at a far distance and leaves sufficient time to react, especially in high-speed driving scenarios.
While being constrained by sensor capability, existing datasets usually contain point clouds of limited ranges, e.g., 120m for KITTI, and 70m for nuScenes \cite{caesar2019nuscenes}, and largely restrict trained algorithms to low-speed driving scenarios. 
When cars drive at 75 mph, the 120m effective range of KITTI and the 70m of nuScenes allow only 3.5s and 2s reaction time, respectively.
Thus, in Cirrus, we adopt LiDAR sensors of a 250-meter effective range, as shown in Figure~\ref{fig:main}, to better support developing and evaluating algorithms for high-speed scenarios.
We present a side-by-side visualization of point clouds from different datasets in a bird-eye view in Figure~\ref{fig:effective}.
Note that, in a point cloud, with respect to distance, the object size stays constant, but the point density varies. Therefore the long effective range of the new dataset provides rich samples with various degrees of point densities, serving a good benchmark for developing algorithms robust across ranges.

In the Cirrus dataset, data are collected using two scanning patterns. 
Besides the standard uniform pattern, the Gaussian pattern gives extra flexibility by enabling sampling with a focus on a particular direction. 
For example, on urban roads, cars drive in a relatively complex environment space, but at a relatively low speed. In this case, the uniform scanning pattern can provide perception from a wide viewing angle at a reduced range. While driving on highways, cars move at a much higher speed, but the environment is considerably simpler. Gaussian scanning pattern can enable longer range detection by focusing on the forward direction, thus allowing for longer reaction time in cases of emergency.
This new dataset enables exploring the merits of each scanning pattern, so that two patterns can be adaptively switched, ideally with a single underlying analysis model through adaptation, based on the driving scenarios for an optimized range-angle trade-off.

The here introduced Cirrus public\footnote{The full dataset will be released upon manuscript acceptance. Examples are available at \url{https://developer.volvocars.com/open-datasets}.} dataset provides fully annotated long-range and bi-pattern paired point clouds,
and enables several potential research topics with great practical impacts.
Based on the aforementioned unique properties, we adopt 3D object detection as a sample application, and perform a series of preliminary experiments on cross-range, cross-device, and cross-pattern adaptations, to illustrate important properties and potential usages of this dataset.
We report results on standard 3D object detection in LiDAR, to serve as a baseline for follow-up studies.
We invite the community to together explore the Cirrus dataset on various tasks for autonomous driving.

Our main contributions are summarized as follows:

\begin{itemize}
	\item We introduce a new long-range bi-pattern LiDAR dataset with exhaustive eight-category object annotations in point clouds over the entire 250-meter sensor effective range.
	
	\item The proposed dataset contains paired point clouds collected simultaneously with Gaussian and uniform scanning patterns, which enables studies on cross-pattern adaptation in point clouds.
	
	\item We adopt 3D object detection in LiDAR as a sample task, and conduct extensive experiments to study model adaptations across ranges, devices, and scanning patterns. Promising results 
	show the great value of this new dataset for future research in the vision and robotics communities.
\end{itemize}

\section{Related Work}
In the past few years, large scale annotated datasets have greatly boosted the research on the perception of the autonomous driving.
Datasets with various sensor setups have been introduced as the development tools for autonomous driving systems.
An RGB camera is the most prevalent sensor thanks to its advantages including the low cost in terms of both hardware and annotations, and the tremendous existing research achievements in computer vision, thus is widely adopted in public datasets \cite{brostow2009semantic,che2019d,cordts2016cityscapes,yu2018bdd100k}.
The rich appearance information in RGB images makes it a suitable choice for inferring semantic.
The works \cite{brostow2009semantic,cordts2016cityscapes} provide high-quality pixel-level  annotations, and are widely adopted in the research of semantic segmentation for autonomous driving. Especially the 5k images with fine annotations and large-scale coarsely annotated samples have paved the way for deep learning based driving support algorithms \cite{chen2017rethinking,long2015fully,noh2015learning}. 
Recently, newly released large scale datasets like BDD100K \cite{yu2018bdd100k} and $D^{2}-$city \cite{che2019d} further enrich the diversity of public datasets by including samples collected under different weathers.
Recent datasets like Apolloscape \cite{huang2018apolloscape} with 144k annotated samples, BDD100K \cite{yu2018bdd100k} with 100k annotated samples, and Mapillary Vistas \cite{neuhold2017mapillary} with 25k samples, significantly enlarge the scale of data for model training. 
However, the significant drawbacks of images-only datasets largely restrict the real-world performance of the image-only systems.
First of all, the inference of distance information from the images is inherently non-trivial and the precision cannot be guaranteed. And the fastly decreased object size in images with respect to the object distance makes it an undesired choice for detecting long-range objects, and therefore is unsuitable for high-speed scenarios where ahead planning is crucial.

To compensate the drawbacks of RGB cameras, object detection with multiple cameras or sensors other than RGB cameras became a popular direction. 
LiDAR is widely adopted for the perception of autonomous driving for the significant advantages including precise localizing and distance measurement, relatively lower noise comparing to RGB images, and the ability to fully perceive multiple angles with a single sensor. 
With the fast development of the deep network based point cloud processing methods \cite{qi2017pointnet,qi2017pointnet++,su2018splatnet} and sparse convolution \cite{liu2015sparse,graham2014spatially,graham2015sparse,graham2017submanifold,graham20183d}, various algorithms \cite{zhou2018voxelnet,shi2019pointrcnn,yan2018second,lang2018pointpillars} are developed to efficiently detect object in point clouds. 
The advantages of the LiDAR sensors and the algorithms make them indispensable components for modern autonomous driving systems, algorithms \cite{ali2018yolo3d,chen2017multi,lang2018pointpillars,li2016vehicle,simon2018complex,yan2018second,zhou2018voxelnet}, and multi-modality datasets \cite{caesar2019nuscenes,chen2018lidar,choi2018kaist,geiger2013vision,patil2019h3d}. 
KITTI \cite{geiger2013vision} provides over 7K annotated samples collected with cameras and LiDAR. 
The stereo images and GPS/IMU data further enables various tasks for autonomous driving.
The KAIST dataset \cite{choi2018kaist} provides data with RGB/thermal camera, RGB stereo, LiDAR, and GPS/IMU, and high diversity with samples collected in both daytime and nighttime. However, the practical value of KAIST is largely restricted by its limited size and 3D annotations.
The intrinsic limitation of LiDAR also cannot be neglected.
Apart from the high cost on hardware, the effective range becomes a bottleneck to the wide adoption of LiDAR in all environments.
Radar sensors, as another popular range sensor, have much longer perception range (250m typically), and lower hardware costs.
But the low point density of radar sensors prevents it from being a qualified replacement to LiDAR. So far only the nuScenes dataset \cite{caesar2019nuscenes} provides point clouds collected with radars.

With the help of modern computer graphics and game engines, synthesized datasets like Playing for Data \cite{richter2016playing}, CARLA \cite{Dosovitskiy17}, Virtual KITTI \cite{Gaidon:Virtual:CVPR2016}, and SYNTHIA \cite{ros2016synthia}, reduce the cost of collecting data, although the performance are sometimes restricted due to the domain shifts between synthesized and real-world data.

\section{The Cirrus Dataset}
The Cirrus dataset contains 6,285 synchronized pairs of RGB, LiDAR Gaussian, and LiDAR uniform frames. All samples are fully annotated for eight object categories across the entire 250-meter LiDAR effective range. 

\subsection{Sensor Placements}

\begin{figure}[]
			\centering
			\includegraphics[width=\linewidth]{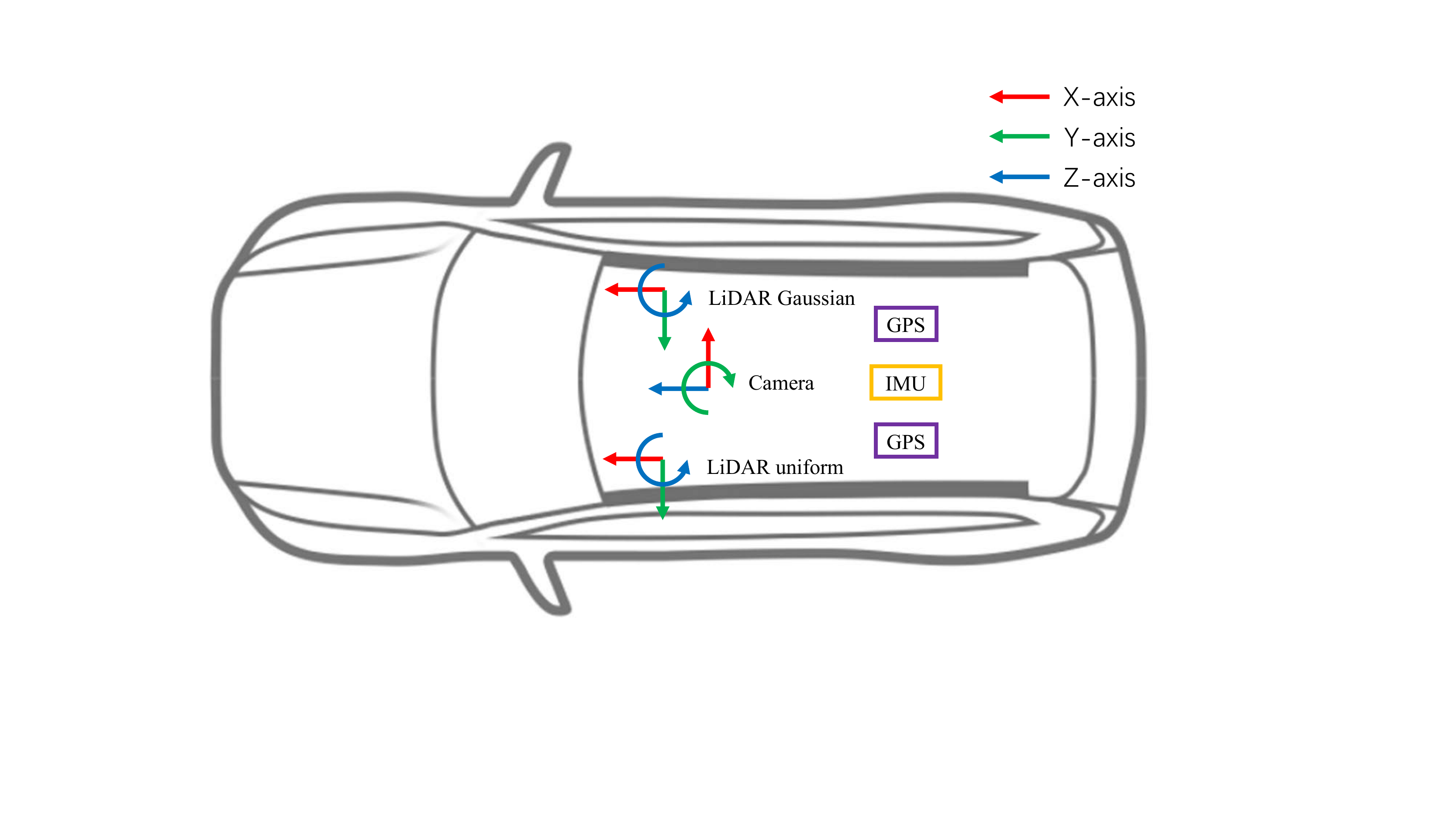}
			\caption{Sensor placements for the dataset collection. All coordination axes follow the right hand rule.}
			\label{fig:setup} 
\end{figure}

The data collection car is equipped with the following sensors:
\begin{itemize}
	\item An RGB camera with a resolution of 1920 $\times$ 650. 
	\item A Luminar Model H2 LiDAR sensor with the Gaussian scanning pattern, 10Hz, 64 lines per frame, 1550-nm, 250m effective range, $>200$ meters range to $10\%$ reflective target (Lambertian), 120$^{o}$ horizontal FOV, 30$^{o}$ vertical FOV. 
	\item A Luminar Model H2 LiDAR sensor with the uniform scanning pattern, 10Hz, 64 lines per frame, 1550-nm, 250m effective range, $>200$ meters range to $10\%$ reflective target (Lambertian),120$^{o}$ horizontal FOV, 30$^{o}$ vertical FOV.
	\item IMU and GPS $\times$ 2.
\end{itemize}
The sensor placements are illustrated in Figure~\ref{fig:setup}.

\subsection{Scanning Patterns}
Two LiDAR sensors mounted on the car are of the identical model, each running a particular scanning pattern. Point clouds are simultaneously captured using both uniform and Gaussian scanning patterns. These two sensors are calibrated to have synchronized and aligned point clouds, and thus annotations can be shared across patterns.

\begin{figure}
	\includegraphics[width=\linewidth]{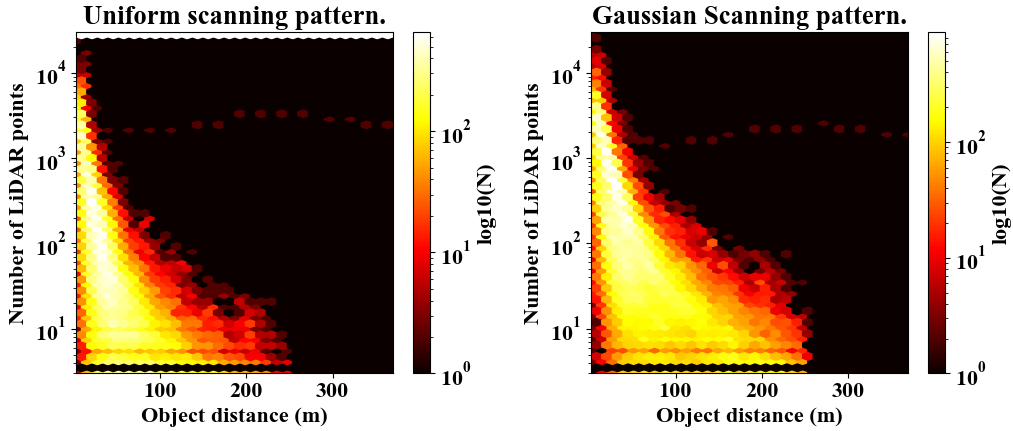}
	\caption{Hexbin log-scaled density plots of the number of LiDAR points inside annotation boxes.
	}
	\label{fig:hexbin} 
\end{figure}

For the LiDAR with the Gaussian scanning pattern, we set the focus of sweeps to the forward direction of the car, which in return, gives higher point density to objects ahead than the uniform pattern.
We plot the Hexbin log-scaled density for both patterns in Figure~\ref{fig:hexbin}.
It is clearly shown in the plot that point clouds collected with the Gaussian pattern have higher overall point density inside annotation boxes.
For long-range objects that are more than 200 meters away, the significantly higher point density in Gaussian pattern point clouds can potentially enable more accurate estimation of object attributes and categories.

\begin{figure}[h]
	\centering
	\includegraphics[width=\linewidth]{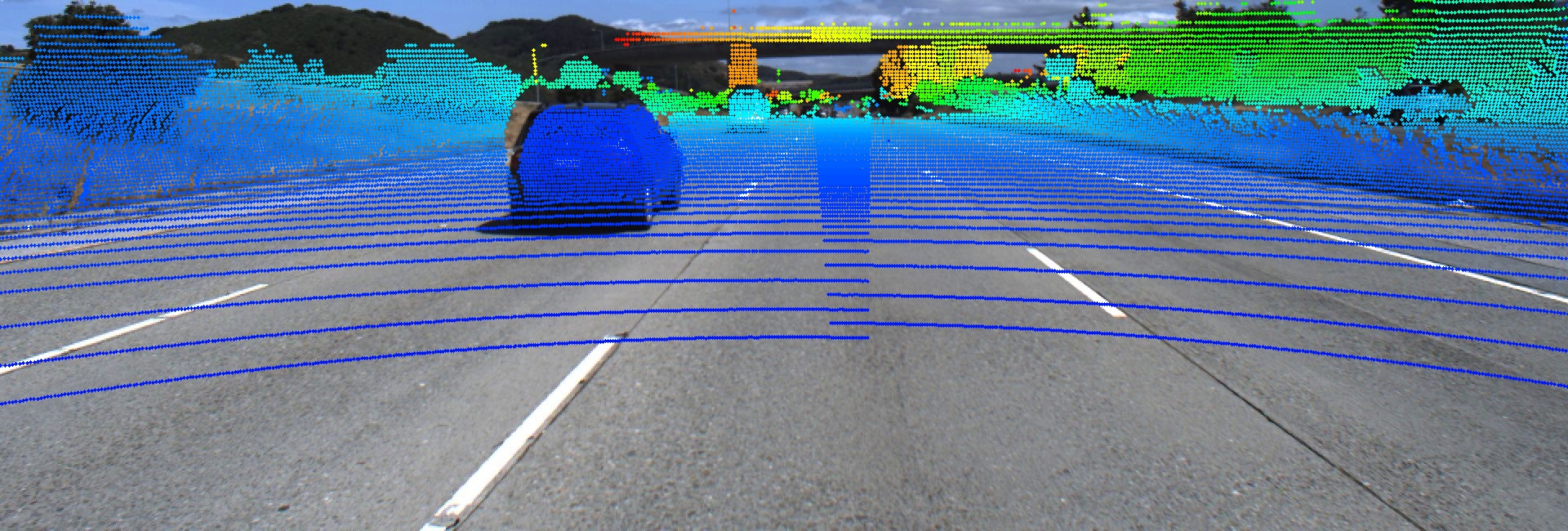}\\
	\vspace{1mm}
	\includegraphics[width=\linewidth]{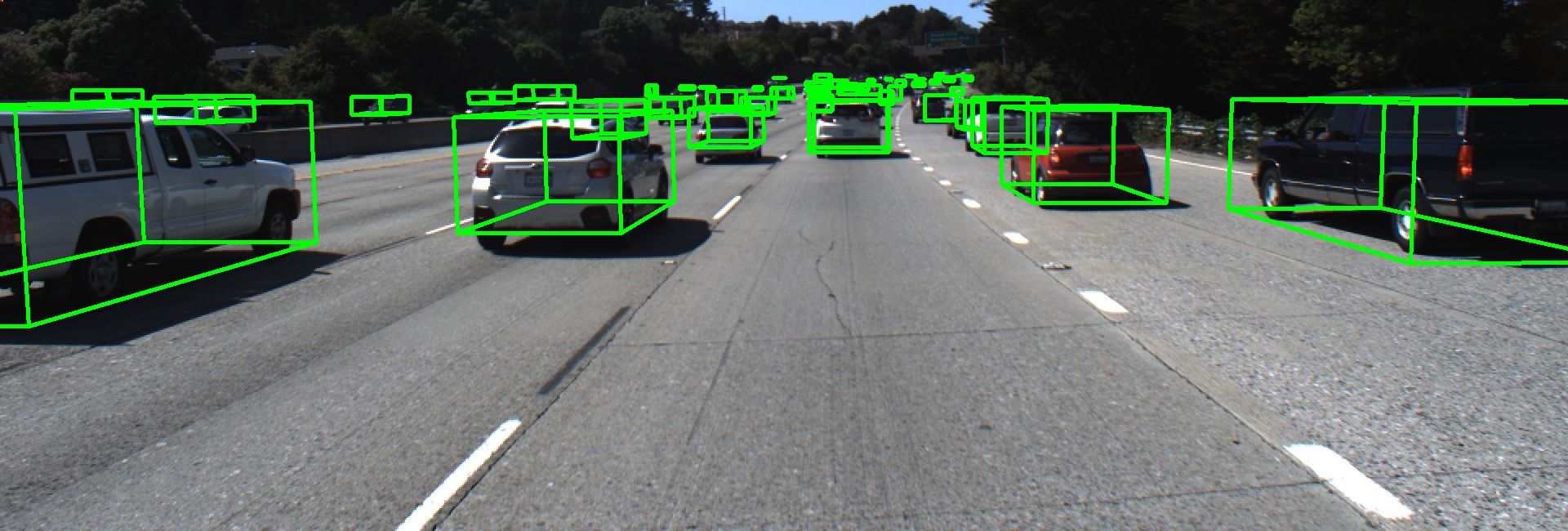}\\
	\vspace{1mm}
	\includegraphics[width=\linewidth]{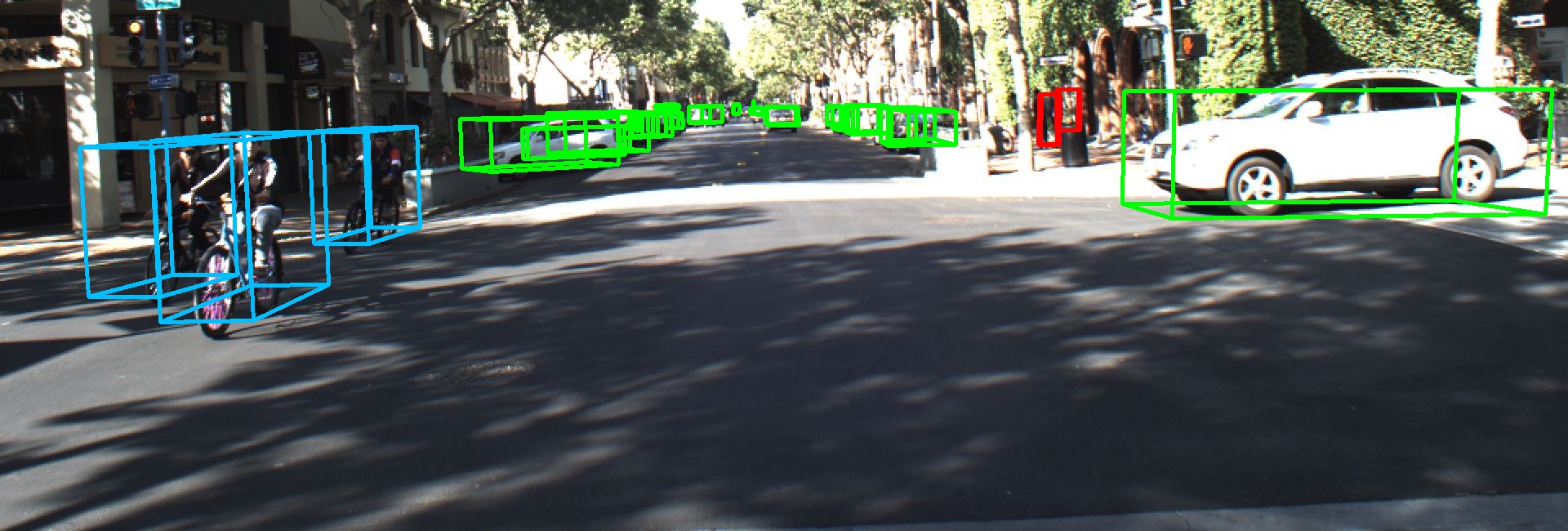}\\
	\vspace{1mm}
	\includegraphics[width=\linewidth]{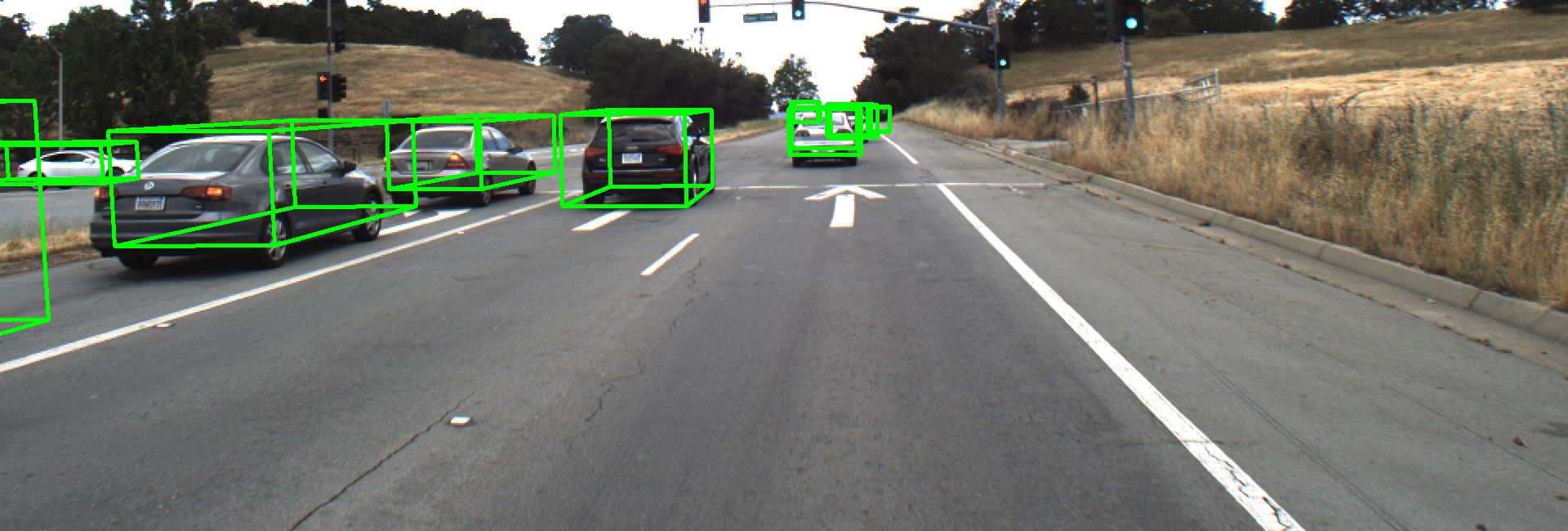}\\
	\caption{Diverse scenes including both highway and urban-road scenarios are included in the Cirrus dataset. We include in the first sample the projected point cloud collected using Gaussian scanning pattern, and then we visualize all the projected boxing boxes in the rest of the samples, where red, blue, and green boxes denote pedestrians, bicycles, and vehicles, respectively. The top-right sample clearly shows the high detection range of the Cirrus dataset where large amount of objects are annotated exhaustively.
	}
	\label{fig:scenes} 
\end{figure}

\subsection{Sensor Synchronization}

The exposure of the camera is triggered first with the corresponding time stamp captured. When the LiDAR sensors start firing and getting returns, the times stamps are generated as well. Then these separate sets of timestamps are sent to the computing platform (such as NVIDIA DRIVE PX2) and the gPTP protocol \cite{ptp} is followed to sync these time stamps. For instance, the camera yields 30 time stamps per second, and each LiDAR sensor gives 10 per second. The nearest matching/syncing across timestamps happens in the PX2. This is a continuous streaming process.

\subsection{Annotations}
Cirrus provides 6,285 frames from 12 segments of videos. 
Both high-speed highway and low-speed urban-road scenarios are included. Example scenes are presented in Figure~\ref{fig:scenes}. 
All images have gone through an anonymization process blurring faces and license plates to eliminate personally identifiable information.

\begin{figure}[t]
	\centering
		\subfigure[Histogram of the number of annotations per category.
		]{
			\includegraphics[width=\linewidth]{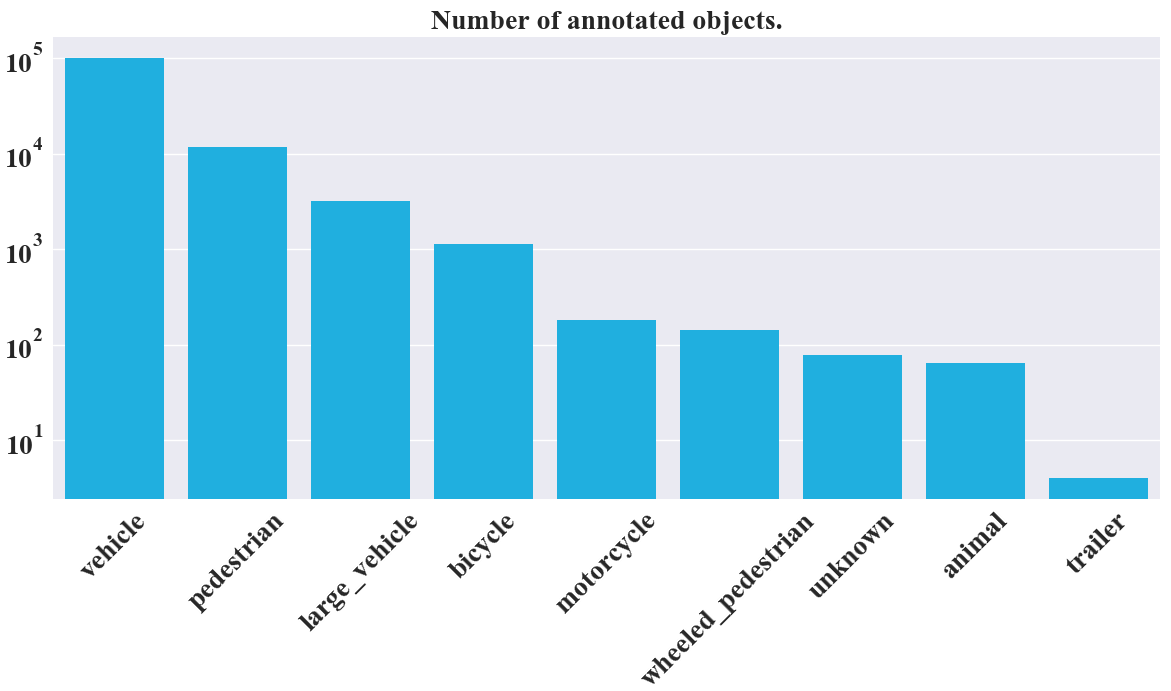}
			\label{fig:hist}
		}
		\hspace{-1mm}
		\subfigure[Histogram of object distance and comparison against KITTI.]{
			\includegraphics[width=\linewidth]{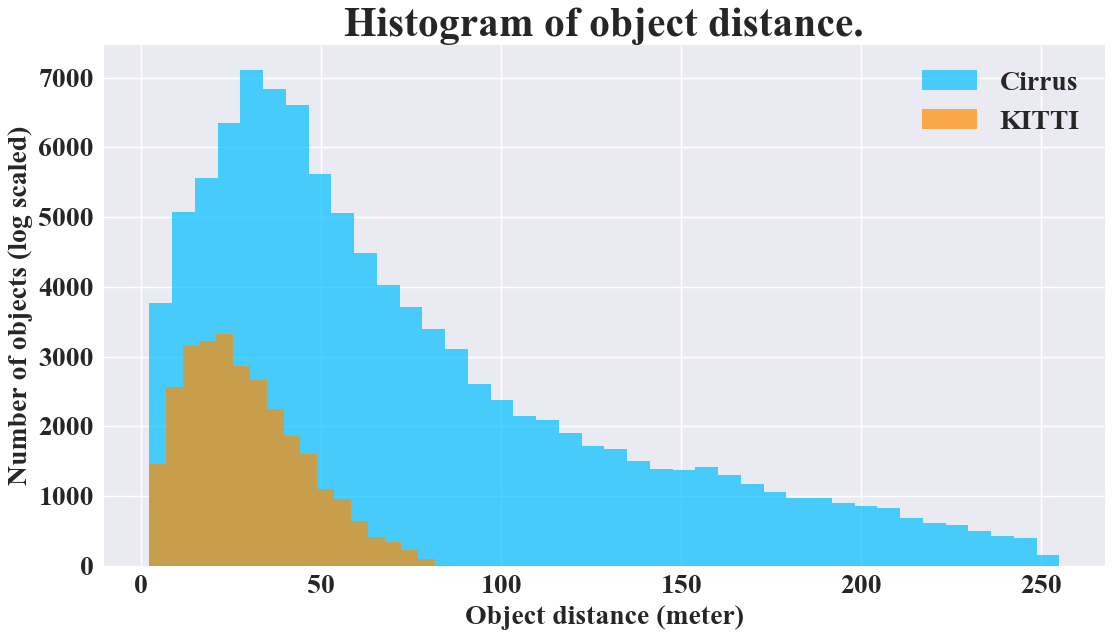}
			\label{fig:dist_hist}
	}
	\caption{Histogram of annotations (top) and comparison with KITTI (bottom). }

\end{figure}

We annotate 8 categories of objects: \textit{vehicle, large vehicle, pedestrian, bicycle, animal, wheeled pedestrian, motorcycle}, and \textit{trailer}.
Objects that do not belong to the aforementioned categories are annotated as \textit{unknown}.
The statistics of the annotated categories are plotted in Figure~\ref{fig:hist}.

For each object from known categories, we annotate its spatial position as $\{x, y, z\}$ in the LiDAR coordinate. 
The shape of each object is represented by its length, width, and height, as well as the rotation angle (yaw) represented using a quaternion.
Different from the previous datasets, where the full body of each object is tightly annotated with a 3D bounding box, the boxes in our dataset contain object parts that are visible in the point clouds.
Since we focus particularly on far-range object detection, and many annotations are beyond visual range, it is hard to infer the full body of every object especially when an object is at a long distance and cannot be seen clearly in the RGB image.

We plot the histogram of object distances for the vehicle (car) category in Figure~\ref{fig:dist_hist}, where it is clearly shown that a large amount of objects appear across the 250-meter effective range. Note that the farthest annotation reaches a distance of over 350 meters.
The histogram comparison against KITTI \cite{geiger2013vision} is also included in Figure~\ref{fig:dist_hist}.
Cirrus provides significantly larger amount of vehicles and objects are widely spread across the longer effective range comparing to KITTI.

\begin{figure*}[h]
	\resizebox{\linewidth}{!}{
		\begin{minipage}{0.8\linewidth}
			\centering
			\captionof{table}{3D object detection with VoxelNet and the Cirrus dataset.}
			\label{tab:bl}
			\begin{tabular}{|c |c| C{18mm} C{18mm} C{18mm} C{18mm}|}
				\hline
				Pattern & Metric & Near range & Mid range & Far range & Overall \\
				\hline
				\hline%
				\multirow{4}*{Gaussian} & DDS & 0.6954 & 0.5594 & 0.3412 & 0.6444 \\
				~&wDDS ($\epsilon = 0.003$) & 0.7112 & 0.5612 & 0.3517 & 0.6546\\
				~&wDDS ($\epsilon = 0.006$) & 0.7232 & 0.5885 & 0.3884 & 0.6630\\
				~&wDDS ($\epsilon = 0.009$) & 0.7421 & 0.5904 & 0.3908 & 0.6691\\				
				\hline
				
				\multirow{2}*{Uniform} & DDS & 0.6297 & 0.4888 & 0.3012 & 0.5672 \\
				~&wDDS ($\epsilon = 0.006$) & 0.6618 & 0.5013 & 0.3234 & 0.5908\\
				
				\hline
			\end{tabular}
		\end{minipage}
		\hspace{1mm}
		\begin{minipage}{0.25\linewidth}
			\centering
			\captionof{table}{3D object detection with state-of-the-art methods. $\epsilon = 0.006$ is set as default for wDDS.}
			\label{tab:more}
			\begin{tabular}{|c | cc| }
				\hline
				Methods & DDS & wDDS \\
				\hline
				\hline%
				VoxelNet & 0.6444 & 0.6630\\
				SECOND & 0.6621 & 0.6819\\
				PointRCNN & 0.6672 & 0.6788 \\
				\hline
			\end{tabular}
		\end{minipage}
		\hspace{4mm}
		\begin{minipage}{0.3\linewidth}
			\centering
			\captionof{table}{Cross-range adaptation. Model adaptation improves the detection performance across the entire effective range.}
			\label{tab:crad} 
			\begin{tabular}{|c |c c |}
				\hline
				Range &  DDS & wDDS \\  
				\hline
				\hline%
				Near range  & 0.7237 & 0.7291  \\
				Far range  & 0.6018 &  0.6098 \\
				Overall & 0.6543 & 0.6825\\
				\hline
				
			\end{tabular}
		\end{minipage}
	}
\end{figure*}

\begin{figure}[h]
	\hspace{-5mm}
	\centering
	\subfigure[Mean Average Precision.]{
		\includegraphics[width=0.5\linewidth]{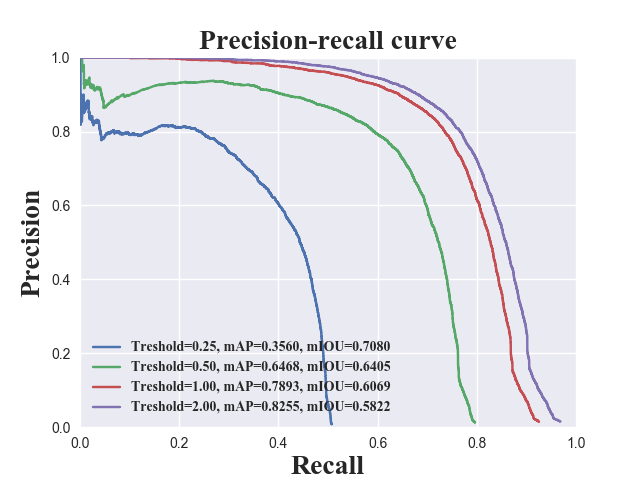}
		\label{fig:map}
	}
	\subfigure[Mean Average Precision with weighted threshold ($\epsilon = 0.006$).]{
		\hspace{-4mm}
		\includegraphics[width=0.5\linewidth]{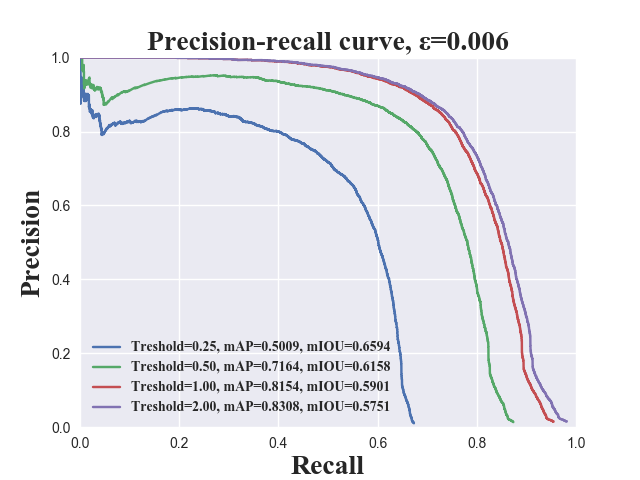}
		\label{fig:wmap6}
	}

	\caption{Precision-recall curves. }
\end{figure}

\section{3D Object Detection and Model Adaptation}
In this section, we present sample tasks, 3D object detection and model adaptation on the newly introduced Cirrus dataset, to illustrate its unique properties and potential usages.
We start this section with evaluation metrics we adopt for this new dataset, and then briefly introduce a benchmark setting for detection and adaptation experiments. All baseline results will be presented in Section~\ref{exp}.

\subsection{Evaluation Metrics}
3D object detection requires accurate estimation of object category, location, and pose.
As pointed out in \cite{caesar2019nuscenes}, 3D intersection over union (IoU) is sensitive to minor drifts for objects with small spatial dimensions such as pedestrians. In our case, since we adopt partial annotations for objects that are not fully visible in the point clouds, we observe unstable assessments for long-range object detection using IoU, which is thus unreliable to faithfully quantify the algorithm performance.
Inspired by the nuScenes detection score (NDS) in \cite{caesar2019nuscenes}, we propose a new decoupled detection score (DDS) that independently evaluates mean average precision for spatial locations, and box attribute estimation for spatial dimensions and poses.
DDS is computed as
\begin{equation}
	\begin{aligned}
		\label{dds}
		\rm{DDS} = \frac{1}{2} ( mAP + aIoU ),
	\end{aligned}
\end{equation}
where mAP and aIoU are mean average precision and aligned intersection over union, respectively, as detailed next.

\noindent{\textit{Mean Average Precision.}}
The partial annotations for the limited-visibility objects including small, occluded, and far distant objects, make the standard IoU based mAP metric over-sensitive to small spatial drifts on the prediction.
In order to decouple the object localization and the box attribution estimation, we use mAP to independently evaluate the precision of object location prediction. 
Specifically, following \cite{caesar2019nuscenes}, a match in average precision is defined by threshholding the 3D spatial distance between the ground-truth and predicted object center $d$ in 3D coordinates. 
AP is calculated as the normalized area under the precision-recall curve. 
The final mAP is calculated by the average over the multiple matching threshold $\mathbb{T} = \{0.5, 1, 2, 4\}$ as
\begin{equation}
	\begin{aligned}
		\label{map}
		\textbf{mAP} = \frac{1}{|\mathbb{T}|}\sum_{t \in \mathbb{T}} \textbf{AP}_t.
	\end{aligned}
\end{equation}

\noindent{\textit{Aligned Intersection over Union.}
With the precision of location prediction measured by mAP, we now introduce aligned intersection over union (aIoU) as the metric to measure the precision of box attribute estimation.
We calculate aIoU as the IoU after aligning the 3D center point of the predicted and the ground-truth object boxes. 
In this way, aIoU only considers the precision of the box shape in terms of both the object dimensions and yaw angle.
The spatial drifts are not included in the calculation of aIoU since they are already measured by mAP.

\noindent{\textit{Weighted Decoupled Detection Score.}
One of the unique properties of our new dataset is the long effective range and the exhaustive annotations across the entire range.
In practice, we consistently observe that the overall performance on the Cirrus dataset is largely constrained by the far-ranges objects which have low-density points and severe occlusions.
To fairly evaluate the performance cross all ranges, we further propose a weighted decoupled detection score (wDDS), which uses a dynamic threshold for objects at different distances.
Specifically, we introduce a new weight factor $\omega$, which is calculated as
\begin{equation}
	\begin{aligned}
		\label{wei}
		\omega = \exp{(\epsilon \cdot d)},
	\end{aligned}
\end{equation}
where $d$ is the distance to an object on a 2D plane, $\epsilon$ is a positive constant, and $\epsilon = 0$ equals to DDS without weighting.
The weight factor $\omega$ is applied by multiplying it with the matching threshold $\mathbb{T}$, so that objects at a far distance will have relatively large thresholds, and the matching thresholds for the objects at a near distance remain close to $\mathbb{T}$.

The proposed DDS and wDDS consider only single object category without taking into account the classification accuracy. When jointly detecting multiple classes of objects, DDS and wDDS for each category are calculated and reported separately.

\begin{figure*}[h]
	\resizebox{\linewidth}{!}{%
		\begin{minipage}{0.42\textwidth}
			\centering
			\captionof{table}{Cross-device adaptation experiments performed on KITTI $\rightarrow$ Cirrus. Different amounts of annotated data in Cirrus are marked in the table.
			}
			\label{tab:cdad} 
			\begin{tabular}{|c |C{13mm} C{13mm}|}
				\hline
				Methods  & DDS  & wDDS \\
				\hline
				\hline
				Pretrained & 0.6506 & 0.6612\\
				Model adaptation (25$\%$) & 0.6511 & 0.6606 \\
				Model adaptation (50$\%$) & 0.6609 & 0.6824 \\
				Model adaptation (full) & 0.6688 & 0.6904 \\
				\hline
			\end{tabular}
		\end{minipage}
		\hspace{3mm}
		\begin{minipage}{0.35\linewidth}
			\centering
			\captionof{table}{Cross-pattern compatibility. G and U denote Gaussian and uniform, respectively.}
			\label{tab:cpjt} 
			\begin{tabular}{|c |C{15mm} C{15mm}|}
				\hline
				Pattern & DDS  &  wDDS\\
				\hline
				\hline%
				G $\rightarrow$ U & 0.5492 & 0.5523  \\
				U $\rightarrow$ G & 0.5514 & 0.5713 \\
				\hline
			\end{tabular}
			
		\end{minipage}
		\hspace{1mm}
		\begin{minipage}{0.44\linewidth}
			\centering
			\captionof{table}{Cross-pattern model performance after joint training and model adaptation.
			}
			\label{tab:cpad} 
			\begin{tabular}{|c |C{12mm}  C{12mm} | C{12mm} C{12mm} |}
				\hline
				Method & \multicolumn{2}{c|}{Joint training} & \multicolumn{2}{c|}{Adaptations} \\
				\hline
				Metric & DDS  & wDDS &  DDS  & wDDS \\
				\hline
				\hline%
				Gaussian & 0.6692 &  0.6770 & 0.6865 & 0.6991\\
				Uniform & 0.5914 & 0.6107 & 0.6057 & 0.6212\\
				\hline
			\end{tabular}
			
		\end{minipage}
	}
\end{figure*}

\section{Experiments}
\label{exp}
In this section, we adopt 3D object detection as a sample application, and perform a series of model adaptation experiments.
\textit{Vehicle} (including \textit{large vehicle}) is selected as the target category to detect. Vehicles dominate the objects annotated in our dataset, and their wide existence across the entire effective range in the current version of our dataset provides reliable assessments to the algorithm performance.
We start with standard 3D object detection and report baseline performance obtained on state-of-the-art 3D object detection methods including VoxelNet \cite{zhou2018voxelnet}, SECOND \cite{yan2018second}, and PointRCNN \cite{pointrcnn}.
Various model adaptation experiments, including cross-device, cross-pattern, and cross-range, are then conducted to validate the value of the proposed benchmark for future research in LiDAR.
VoxelNet \cite{zhou2018voxelnet} provides a principle way of efficient object detection in point clouds, and is used as the baseline network for performing adaptation experiments. And a Gaussian scanning pattern is selected as the default pattern besides the cross-pattern adaptation, where data for both patterns are used.

\subsection{3D Object Detection}
We start with standard 3D object detection using the new Cirrus dataset.
We train VoxelNet to detect vehicles represented by bounding boxes with 3D location, 3D dimension, and yaw angle in LiDAR.
To produce gridded features for convolutional layers, following \cite{zhou2018voxelnet}, we convert the point clouds into equally spaced 3D voxels.
The detection range is set to be $[0, 250] \times [-50, 50] \times [-3, 1]$ meters along the X, Y, Z axis, respectively. 
The voxel size is set to be vW = 0.2, vH = 0.2, vD = 0.4 meters,
which leads to gridded feature maps with a size of $1250 \times 500 \times 10$ that allow accurate box location estimation at high resolution feature maps.

Models for point clouds with Gaussian and uniform scanning patterns are trained separately.
The results are presented in Table~\ref{tab:bl}.
To comprehensively evaluate the algorithm robustness across range, we divide the 250 meter effective range into three levels: 0-70 meters as the near range, 70-150 meter as the mid range, and 150-250 meters as the far range.
The performance for each range is reported separately, followed by an overall performance across the entire detection range.
We report performance measured by both DDS and wDDS with 4 values of $\epsilon$, and the precision-recall curves with $\epsilon = 0$ and $\epsilon = 0.006$ are plotted in Figure~\ref{fig:map} and Figure~\ref{fig:wmap6}. We select $\epsilon = 0.006$ as the default setting of wDDS in the following experiments.

We further provide results on more state-of-the-art methods in Table~\ref{tab:more} for benchmarking future methods.

\subsection{LiDAR Model Adaptations}
\noindent{\textbf{\textit{Cross-range Adaptation}}
We firstly show that, for a long effective range, we can improve the overall algorithm robustness by encouraging consistent deep features across the entire range. 
We adopt the framework of range adaptation proposed in our previous work \cite{Wang_2019_ICCV} for promoting consistent feature across range both locally and globally.
We use point clouds with Gaussian pattern, and divide the 250-meter effective range into two areas, with 0-100 meter as the near range and 100-250m as the far range to perform cross-range adaptation from near range to far range.
The results are presented in Table~\ref{tab:crad}; we report DDS and wDDS after adaptation on near-range, mid-range, and far-range areas.
The performance improvements indicates that the invariant feature benefits object detection across the entire effective range.

\noindent{\textbf{\textit{Cross-device Adaptation.}}}
We now consider a more challenging setting, where the cross-domain data is collected by different sensor models.
We adopt point clouds in KITTI, which are collected using LiDAR sensor with shorter effective range and uniform scanning pattern, and perform the cross-device adaptation experiments.
To further validate the practical value of model adaptation against insufficient annotated data, we progressively remove annotated data from Cirrus to show the model adaptation performance with insufficient annotated data.
Note that different from the previous two adaptation experiments, where the network parameters are shared across domains completely, we train domain-specific detection heads for each domain due to the difference on annotation protocol (full-body annotation for KITTI and partially annotation for Cirrus).
The results are presented in Table~\ref{tab:cdad}.
We also present results on training the network using reduced amount of data from Cirrus alone to show the performance improvement with model adaptation.

\noindent{\textbf{\textit{Cross-pattern Adaptation.}}}
In this experiment, we perform model adaptation across the Gaussian and the uniform scanning patterns, so that one common model supports dynamic switching between different scanning patterns.
We start with directly feeding a model trained using one scanning pattern with point clouds from the other pattern. 
As shown in Table~\ref{tab:cpjt}, accuracies drop for point clouds collected with different scanning pattern compared to overall accuracies in Table~\ref{tab:bl}, which indicates that the two scanning patterns are inherently different and the model cannot be shared across patterns directly.
Based on the aforementioned empirical observations, we perform model adaptation with both paired and unpaired cross-pattern point clouds.

\noindent{\textbf{\textit{Cross-pattern Adaptation with Paired Data.}}}
Since we collect our dataset using two LiDAR sensors with different scanning patters simultaneously, and the coordinates are well-calibrated, we have paired data with consistent annotations.
For cross-pattern adaptation with paired data, we directly feed the network with paired point clouds with two patterns and minimize the distance between two features. Feature extractor and detection heads are shared across two patterns.
The results are presented in Table~\ref{tab:cpad} as joint training.

\noindent{\textbf{\textit{Cross-pattern Adaptation with Unpaired Data.}}}
Paired cross-domain data is expensive to collect. In practice, unpaired data is usually more accessible.
In this experiment, we manually shuffle the data collected using Gaussian and uniform scanning patterns, and adopt the adaptation framework to encourage invariant features for both patterns.
The results are presented in Table~\ref{tab:cpad} as adaptations.
Training the network with model adaptation outperforms joint training, indicating the explicit invariant feature imposed by model adaptation improves the generalization of deep networks to different scanning patterns.

\section{Conclusion}
In this paper, we introduced Cirrus, a new long-range bi-pattern LiDAR dataset for autonomous driving.
The new dataset significantly enriches the diversity of public LiDAR datasets by providing point clouds with 250-meter effective range, as well as Gaussian and uniform scanning patterns.
We presented details on the dataset collection and object annotation.
3D object detection in LiDAR is presented as an example task using the Cirrus dataset, and various model adaptation experiments are performed to illustrate important properties and sample usages of this new public dataset.

\bibliographystyle{ieee_fullname}
\bibliography{egbib}

\end{document}